\documentclass[letterpaper, 10pt, conference]{ieeeconf}  

\IEEEoverridecommandlockouts                              
\usepackage{macros-antoine}
\usepackage[color=white!80!black,textsize=scriptsize]{todonotes}
\usepackage{hyperref}
\usepackage{rotating}
\usepackage{acronym}
\usepackage{comment}
\let\labelindent\relax
\usepackage{enumitem}
\usepackage[font=small]{caption}
\acrodef{BO}[BO]{Bayesian Optimization}
\acrodef{CMAES}[CMA-ES]{Covariance Matrix Adaptation Evolution Strategy}
\acrodef{GCS}[GCS]{Graphs of Convex Sets}
\acrodef{RO}[RO]{range-only}
\acrodef{SDP}[SDP]{Semidefinite Program}
\acrodef{SOS}[SOS]{Sums of Squares}
\acrodef{TO}[TO]{Trajectory Optimization}
\acrodef{PL}[PL]{Policy Learning}
\acrodef{iLQR}{Iterative Linear Quadratic Regulator}
\acrodef{FDDP}{Feasibility-driven Differential Dynamic Programming}
\acrodef{RL}{Reinforcement Learning}
\acrodef{TD3}{Twin Delayed DDPG}
\acrodef{PPO}{Proximal Policy Optimization}
\acrodef{POP}{Polynomial Optimization}
\acrodef{RKHS}{Reproducing Kernel Hilbert Space}
\acrodef{SLAM}{Simultaneous Localization and Mapping}

\usepackage[
    backend=biber,
    style=ieee,
    doi=false,
    url=false,
]{biblatex}
\AtEveryBibitem{
  \clearfield{note}
  \clearfield{eprint}
}

\usepackage{bm}

\newcommand{\kernelSOS}{\textit{KernelSOS}}
\newcommand{\ie}{\emph{i.e.}}

\newcommand{\highlight}[1]{\textcolor{red}{#1}}
\renewcommand{\highlight}[1]{#1}

\definecolor{tab10-blue}{RGB}{31,119,180}
\definecolor{tab10-orange}{RGB}{255,127,14} 
\definecolor{tab10-green}{RGB}{44,160,44} 
\definecolor{tab10-red}{RGB}{214,39,40} 
\definecolor{tab10-purple}{RGB}{148,103,189} 
\definecolor{tab10-5}{RGB}{140,86,75}
\definecolor{tab10-6}{RGB}{227,119,194}
\definecolor{tab10-7}{RGB}{127,127,127}
\definecolor{tab10-8}{RGB}{188,189,34}
\definecolor{tab10-9}{RGB}{23,190,207}

\title{\LARGE \bf
Sampling-Based Global Optimal Control and Estimation \\ via Semidefinite Programming
}

\author{Antoine Groudiev\textsuperscript{1}, Fabian Schramm\textsuperscript{1,2}, Éloïse Berthier\textsuperscript{3}, Justin Carpentier\textsuperscript{1,2},
and Frederike Dümbgen\textsuperscript{1,2}%
\thanks{\textsuperscript{1}École Normale Supérieure, PSL Research University, Paris, France.
{\tt\small firstname.lastname@ens.psl.eu}
}
\thanks{\textsuperscript{2}Inria, Département d'informatique de l'ENS, CNRS, PSL Research University, Paris, France.
{\tt\small firstname.lastname@inria.fr}
}%
\thanks{\textsuperscript{3}U2IS, ENSTA, Institut Polytechnique de Paris, Palaiseau, France.
{\tt\small firstname.lastname@ensta.fr}}
\thanks{
This work has received support from the French government, managed by the National Research Agency, through the NIMBLE project (ANR-22-CE33-0008) and under the France 2030 program with the references Organic Robotics Program (PEPR O2R) and “PR[AI]RIE-PSAI” (ANR-23-IACL-0008). It was also funded by the European Union, through the Horizon Europe research and innovation program under the Marie Skłodowska-Curie (GA no.101207106), the ARTIFACT project (GA no.101165695), and through the AGIMUS project (GA no.101070165). Paris Île-de-France Région supported this work in the framework of DIM AI4IDF. 
Views and opinions expressed are those of the author(s) only and do not necessarily reflect those of the funding agencies.
}
}

\begin{document}

\maketitle

\begin{abstract}
Global optimization has gained attention over the past decades, thanks to the development of both theoretical foundations and efficient numerical routines. 
Among recent advances, Kernel Sum of Squares (\kernelSOS) provides a powerful theoretical framework, combining the expressivity of kernel methods with the guarantees of SOS optimization.
In this paper, we take \kernelSOS{} from theory to practice and demonstrate its use on challenging control and robotics problems. 
We identify and address the practical considerations required to make the method work in applied settings: restarting strategies, systematic calibration of hyperparameters, methods for recovering minimizers, and the combination with fast local solvers.
As a proof of concept, the application of \kernelSOS{} to robot localization highlights its competitiveness with existing SOS approaches that rely on heuristics and handcrafted reformulations to render the problem polynomial. Even in the high-dimensional, non-parametric setting of trajectory optimization with simulators treated as black boxes, we demonstrate how \kernelSOS{} can be combined with fast local solvers to uncover higher-quality solutions without compromising overall runtimes.
\end{abstract}

\section{Introduction}

Many problems in robotics and automatic control are phrased as optimization instances, including estimation, control, and planning. 
For many applications, the optimization problems are non-convex, and commonly used local solvers are likely to converge to local optima rather than global ones. This may lead to \textit{inefficiencies} when the system pursues suboptimal plans, and to \textit{safety} problems when it relies on inaccurate state estimates.  
To ensure the quality of solutions for \textit{safe and efficient} operation, methods based on \ac{SOS} and the moment hierarchy have been adopted in robotics and control over the past decades. 
These methods employ convex relaxations of the original non-tractable optimization problems to identify globally optimal solutions~\cite{lasserre,parrilo_semidefinite_2003}.

\ac{SOS} methods suffer from limited generalizability as they are restricted to optimization problems in parametric, polynomial form.\footnote{Some finite-dimensional, non-polynomial basis functions, such as trigonometric functions, have been discussed in the literature~\cite{lofberg_coefficients_2004}, but are still not widely used to this date.} This restriction is problematic in at least two ways. First, many problems of practical interest are not inherently polynomial. As a result, various workaround techniques have been proposed to force a polynomial structure, for example, for distance-based~\cite{beck_exact_2008,halsted_riemannian_2022} and rotation-based~\cite{barfoot_certifiably_2024} estimation problems. 
While often effective in practice, such reformulations may slightly deviate from the desired cost function and may require extensive domain expertise and creativity to discover. Second, scalability issues may arise when a high degree of the moment hierarchy is needed to achieve sufficient \emph{relaxation tightness}, a prerequisite to retrieve certifiably optimal solutions~\cite{yang_teaser_2020,khadir_piecewise-linear_2021}.
\begin{figure}[t]
    \centering
    \includegraphics[width=\linewidth]{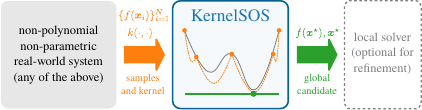}
    \caption{\textbf{Overview of the approach.} 
    This paper demonstrates the effectiveness of \kernelSOS~for optimization problems in estimation and control. Based on function evaluations only, \kernelSOS~can address non-polynomial and non-parametric problems beyond the reach of classic Sum of Squares (SOS) methods. Each iteration of \kernelSOS~solves a single semidefinite program, and multiple restarts yield an approximately globally optimal solution, which can be refined by a local solver.}\label{fig:high-level-method}
\end{figure}

Kernel Sum of Squares (\kernelSOS), a global optimization method recently introduced by Rudi et al.~\cite{rudi_finding_2024}, offers the theoretical and algorithmic foundations to overcome these limitations. While originally studied in a theoretical context and applied in domains such as optimal transport~\cite{vacher24OTKernelSOS}, its potential in optimal estimation and optimal control remains largely underexplored.

The high-level idea of \kernelSOS{} is illustrated in Fig.~\ref{fig:high-level-method}. Intuitively, it replaces polynomial bases with kernel functions, and formulates the optimization problem using only a finite number of function and kernel evaluations. This provides several benefits: 
 (i)~this allows us to consider a larger class of feature functions, such as those stemming from Gaussian or Laplace kernels, thereby addressing the limited expressivity of polynomial features;  
 (ii)~the size of the resulting optimization problem is defined by the number of samples, making the method not directly dependent on the problem dimension, and more amenable to account for limited computing resources;
(iii)~because \kernelSOS{} relies solely on function evaluations, it can address not only non-polynomial objectives but even problems with no explicit parametric form, making it amenable to data-driven rather than model-based settings. 

In this paper, we bring \kernelSOS{} from theory to practice by demonstrating its practical application in optimal estimation and optimal control, two representative and fundamental problems in automatic control theory and robotics. In particular, we make the following contributions:
\begin{itemize}[leftmargin=*]
    \item As a proof of concept, we demonstrate that \kernelSOS{} can solve \ac{RO} localization, where classic \ac{SOS} approaches require heuristics~\cite{beck_exact_2008,dumbgen_safe_2023}, or a significant increase in problem dimension~\cite{halsted_riemannian_2022}. 
\item Next, we show how \kernelSOS{} can seamlessly integrate complex components, such as high-fidelity dynamics simulators, into the solution process of \ac{TO}, outperforming state-of-the-art zeroth-order solvers~\cite{hansen_cma_2023} in low-sample regimes.
\item Finally, we demonstrate the effectiveness of combining \kernelSOS{} with local solvers: global search from \kernelSOS{} uncovers promising regions of the solution space, while local methods refine them to high accuracy. This combination strikes a balance between global exploration and local efficiency, offering a practical approach to tackling challenging control and robotics problems.
\end{itemize}
\section{Related Work}\label{sec:related}

\subsection{Sum of Squares methods}

\ac{SOS} and moment-hierarchy methods have been used for a variety of tasks in control and estimation. In control, this includes region-of-attraction estimation~\cite{henrion2013convex}, control synthesis using funnel methods~\cite{tedrake2010lqr}, globally optimal control~\cite{lasserre2008nonlinear,berthier2022infinite}, or inverse kinematics~\cite{giamou_convex_2022}. In estimation, problems like robust estimation~\cite{yang_certifiably_2023}, rotation averaging~\cite{eriksson_rotation_2018}, or simultaneous localization and mapping~\cite{rosen_se-sync_2019,carlone_duality-based_2015} can be solved to certifiable global optimality thanks to this framework.

\kernelSOS~can be seen as a generalization of (polynomial) \ac{SOS} methods (and, by duality, moment hierarchy methods) for general feature vectors. 
 In particular, it extends naturally to infinite-dimensional basis vectors. While still in its early stages, it was already applied to optimal control problems~\cite{berthier2022infinite}, to learn value functions, using a linear-programming formulation similar to that in~\cite{lasserre2008nonlinear}. Here, we explore a broader and still under-studied application of the \kernelSOS{} method in control and estimation: its potential to solve general non-polynomial and non-parametric optimization problems.

\subsection{Non-parametric optimization}
Non-parametric methods (sometimes called \textit{black-box} methods) only require feasible samples to solve an optimization problem. Prominent examples include evolutionary algorithms such as \ac{CMAES}, and sequential approaches such as \ac{BO}. \ac{CMAES}~\cite{hansen_cma_2023} evolves a \emph{population} of samples over multiple \emph{generations}, alternating between sampling from a model distribution, usually a multivariate Gaussian, and then updating the distribution based on a subset of the new samples. 
\ac{BO}~\cite{kushner_new_1964}, on the other hand, aims to find a surrogate model by iteratively choosing optimal samples by maximizing an acquisition function, seeking, for example, the maximum expected improvement. The acquisition function and final surrogate functions are, however, although non-convex, commonly optimized using local solvers.

Just as \ac{CMAES}, \kernelSOS{} also evolves a population of samples over multiple iterations, but with a global minimization performed per iteration. Rather than using a local Gaussian, it uses a global surrogate model as in \ac{BO}, but with random rather than active sampling, and fits and globally minimizes the surrogate function in one shot using a~\ac{SDP}.  

\section{KernelSOS for Global Optimization}\label{sec:method}
\begin{figure}[t]
    \centering
    \includegraphics[width=\linewidth]{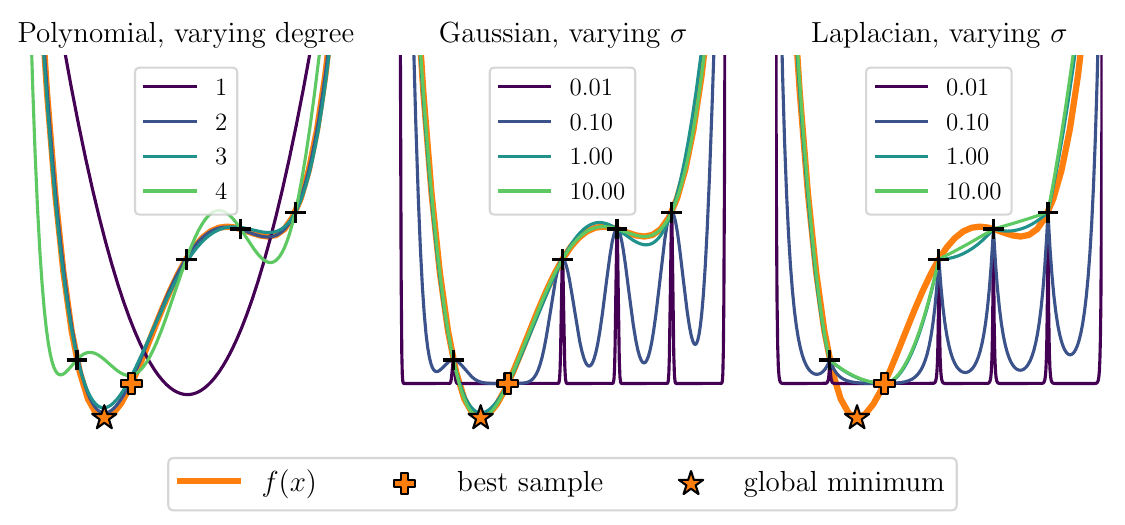}
    \caption{\textbf{Illustration of \kernelSOS{} with different kernels}. 
    Given samples (black crosses) of an arbitrary (possibly unknown) function $f$ (orange), \kernelSOS~minimizes a kernel-defined surrogate function (blue tones), by solving a simple \ac{SDP}. When the kernel is chosen so that $f$ lies in its \ac{RKHS}, a good estimate of the global minimum is found, even for a low number of samples. Here, $f$ is a polynomial of 4th degree, and therefore the polynomial kernel of degrees 2 and 3, and the Gaussian kernel with $\sigma>1$, lead to good results. The width of the plot is 2.}\label{fig:kernelsos-visualization}
    \label{fig:surrogates}
\end{figure}

This section gives a high-level overview of the \kernelSOS~method, as introduced in~\cite{rudi_finding_2024}. 
We are interested in optimization problems of the form
\begin{equation}\label{eq:minf}
    \min_{\bm{x}\in\Omega} f(\bm{x}),
\end{equation}
where $f$ is either unknown or too complex to be treated by standard solvers, and $\Omega$ is a compact and convex set\footnote{While the original paper~\cite{rudi_finding_2024} has more general assumptions, we restrict ourselves to such sets for the implementation; all search spaces considered in the following satisfy these assumptions.} on which we can sample $f$. All information available to the solver is evaluations of the function $f$ at a finite number of points $\bm{x}_i$, $i\in[N]$, with shorthand notation $[N]:=\{1,\ldots, N\}$.

\subsection{SOS formulation}
It is useful to recall the standard (equations-based) \ac{SOS} framework, a powerful tool to find global minimizers of polynomial functions.
Assuming $f$ attains its global minimizer, the following problem is equivalent to~\eqref{eq:minf}: 
\begin{equation}\label{eq:maxc}
    c^* = \max_{c\in \R} \, c \quad \text{ s.t. } \forall \bm{x}\in\Omega: f(\bm{x}) - c \geq 0 \, .
\end{equation}
Problem~\eqref{eq:maxc} finds the largest lower bound $c$ of the function~$f$. The problem is convex but has infinitely many constraints.  Evaluating if a function (here, $f(\bm{x}) - c$) is non-negative is, in general, NP-hard. However, a sufficient (but generally non-necessary) condition for non-negativity of a polynomial is that it can be written as a sum of other polynomials squared; such polynomials are called \ac{SOS} and can be written as $p(\bm{x})=\bm{v}_d(\bm{x})^\top \bm{A} \bm{v}_d(\bm{x})$, with $\bm{v}_d$ a polynomial basis vector of degree $d$ and $\bm{A}\succeq 0$. 
Using this restriction, we obtain:
\begin{equation}\label{eq:maxc-sos}
\begin{aligned}
    c_{\text{d-SOS}}^* = \max_{c\in \R, \bm{A}\succeq 0} \,  c ~~& \\ 
    \text{ s.t. }   f(\bm{x}) & - c  = {\bm{v}_d(\bm{x})}^\top \bm{A} \bm{v}_d(\bm{x}),\quad\bm{x}\in\Omega \,.
\end{aligned}
\end{equation}
Because we have replaced Problem~\eqref{eq:maxc} by the~\ac{SDP} \textit{restriction}~\eqref{eq:maxc-sos}, we have $c_{\text{d-SOS}}^*\leq c^*$. Since some non-negative polynomials are not \ac{SOS}, $c_{\text{d-SOS}}^*=c^*$ does not hold in general. The difference between the optima is also known as the \textit{relaxation gap}, 
and when it is zero, we have a \textit{tight} relaxation. Under mild assumptions, the gap is guaranteed to approach zero as we increase the degree~$d$~\cite{parrilo_semidefinite_2003,nie_optimality_2014}.
Sometimes, tightness is achieved at a low degree\footnote{\highlight{More specifically, finite convergence is generic for polynomial optimization problems~\cite{nie_optimality_2014}.  However, knowing whether finite convergence will appear for a given instance is NP-hard~\cite{vargas_hardness_2024}.}}; however, in many scenarios, the problem becomes prohibitively expensive before a sufficient level of tightness is attained. 

\subsection{KernelSOS formulation}\label{sec:ksos-formulation}

The \kernelSOS~formulation replaces the basis functions $\bm{v}_d(\bm{x})$ (often the monomial basis), by possibly infinite-dimensional functions in a chosen~\ac{RKHS}. The \ac{RKHS} and this class of functions are uniquely identified by choosing a positive definite kernel~\cite{vert2004primer}, which can be significantly easier than providing the functions in parametric form.\footnote{
Choosing a polynomial kernel recovers the classic \ac{SOS} formulation~\cite{rudi_finding_2024}.} 

Using this kernelized form, we can formulate the problem in terms of pairwise kernel evaluations $k(\bm{x}_i,\bm{x}_j)$ and cost evaluations $f(\bm{x}_i)$ at a finite number of samples $\bm{x}_i$. This means we can include non-parametric elements, such as a simulator or real-world runs.
We define the kernel matrix $\bm{K}={\left(k(\bm{x}_i,\bm{x}_j)\right)}_{i,j \in [N]}$, where $k$ is a chosen positive-definite kernel. We denote its Cholesky decomposition by $\bm{K}=\bm{R}^\top\bm{R}$, with $\bm{R}$ an upper-triangular matrix. Using a dedicated representer theorem~\cite{marteau_non-parametric_2020}, the following finite-dimensional~\ac{SDP} can be derived:
\begin{align}\label{eq:max-ksos}
    \begin{split}
    c_{\text{k-SOS}}^* = \max_{c\in \R,\bm{B}\succeq 0} \, &c - \lambda\text{Tr}({\bm{B}})\\ 
    \text{ s.t. } & f(\bm{x}_i) - c = \bm{\Phi}_i^\top \bm{B} \bm{\Phi}_i, \,\, i\in[N]\,,
    \end{split}
\end{align}
where the $\bm{\Phi}_i$ are the columns of $\bm{R}$, $\text{Tr}$ is the trace operator, and $\lambda\geq0$ is a regularization parameter related to the finite number of samples $N$ and the smoothness of $f$ with respect to the kernel~$k$. 

Choosing appropriate values for the kernel, $\lambda$, and $N$, is crucial.
In Fig.~\ref{fig:kernelsos-visualization}, generated with $\lambda=0$ for the polynomial kernel, and $\lambda=10^{-3}$ for the exponential kernels, we visualize a univariate polynomial example problem. Alongside $f$, we plot what we call the \emph{surrogate function},\footnote{This is called the \emph{model} in the original work and can be computed using~\cite[Equation (37)]{rudi_finding_2024}.} which corresponds to the optimal function in the chosen \ac{RKHS} that passes through the given samples. Without careful calibration of the parameters, the surrogate function is a poor match of $f$, and the solution is equally poor. This calibration, along with other ingredients for successfully applying the \kernelSOS{} algorithm to robotics and control applications, is the subject of the next section. 
\section{KernelSOS in Practice}
\label{sec:ksos-in-practice}
In this section, we provide methodological insights to bring \kernelSOS{} from theory to practice. We identify and address the practical ingredients needed for its successful application to high-dimensional, non-parametric problem formulations.

\subsection{Finding the minimizer}

In~\cite{rudi_finding_2024}, two methods to recover a good candidate of $\bm{x}^\star$, the minimizer of $f$, are discussed. First, one can use:
\begin{equation}\label{eq:sol}
    \bm{x}^\star = \sum_{i=1}^N \alpha_i^\star \bm{x}_i,
\end{equation}
where $\alpha_i^\star$ for $i \in [N]$ are the optimal dual variables corresponding to the equality constraints in~\eqref{eq:max-ksos}. We provide a new intuition for this candidate, using the dual problem, in Appendix~\ref{sec:appendix-dual}. This candidate is exact for sufficiently many samples, depending on the smoothness of $f$. An alternative method that maximizes a lower-bounding parabola instead of the constant $c$, is also proposed in~\cite[Sec.~7.1]{rudi_finding_2024}. In practice, we found that this latter method exhibits poor conditioning: low curvature is required for the solution to be viable, but this leads to poor sensitivity at the optimum.

\subsection{Restarting strategy}

The runtime of solving the semidefinite program~\eqref{eq:max-ksos} scales cubically with the number of samples~$N$.
To improve scalability, we implement the restarting procedure mentioned in~\cite{rudi_finding_2024} as a byproduct to improve convergence rates, which iteratively shrinks the region to sample from. 
The algorithm is given an initial center $\bm{z}_0$ and a radius $r_0$. At each iteration, the algorithm solves the optimization problem with $N$ new samples in the region $B(\bm{z}_i, r_i)$, where $B(\bm{z}_i, r_i)$ is the hypercube of radius $r_i$ centered at $\bm{z}_i$. The algorithm then sets $\bm{z}_{i+1}$ to the minimizer of iteration $i$, and $r_{i+1}=\gamma r_i$, where $\gamma\in(0,1)$ is a decay factor. We denote the number of such \emph{restarts} by $w$.

\subsection{Choosing hyperparameters}\label{sec:hyper}

Hyperparameters of the \kernelSOS~method include the number of samples $N$, the number of restarts~$w$, the shrinking factor $\gamma$, the kernel type and its parameters, and the regularization factor~$\lambda$.
For good performance of the method, it is essential to understand how these parameters interact. 

\noindent
\textbf{Kernel and scale.} The choice of kernel and its scale $\sigma$ should reflect the smoothness of the objective function $f$, since they determine the function class (\ie, the \ac{RKHS}) that we restrict the optimization to.  
In practice, less smooth functions require denser sampling to capture their variations, together with a smaller~$\sigma$.
A good rule of thumb is to choose $\sigma$ such that it has the same order of magnitude as the minimum distance between samples. Accordingly, in the restarting scheme, we reduce the kernel width as the search region shrinks, following the update rule $\sigma_{r+1}=\gamma\sigma_r$.

In Figure~\ref{fig:surrogates}, we show the surrogate functions obtained for different choices of kernels, for a quartic univariate optimization problem. The polynomial kernel is highly sensitive to the correct choice of the degree.\footnote{For the polynomial kernel, to avoid infeasibility, the equality constraints are relaxed to soft constraints. We add the term $\frac{\mu}{2}\|\bm{r}\|^2$ to the cost, where $\bm{r}$ are the constraint residuals, and estimate the dual variables using $\bm{\alpha}\approx\mu \bm{r}$.} The Gaussian and Laplacian represent poor approximations at low $\sigma$ values, and with the Gaussian, which represents smoother functions than the Laplacian kernel, fitting the function particularly well as $\sigma$ approaches 10.

\noindent
\textbf{Regularization parameter.} The parameter $\lambda$ influences the bias-variance tradeoff of the optimization, as in (kernel) ridge regression~\cite[Chapter 3.6]{bach_learning_2024}. Keeping all other parameters fixed, increasing~$\lambda$ favors more regular functions of the RKHS, as it is related to its effective dimension. This decreases the variance of the estimation — more regular functions being easier to estimate — but increases the bias, as we reduce the space of candidate functions. 

\noindent\textbf{Asymptotic properties.}  
It is useful to understand the asymptotic properties of $\sigma$ and $\lambda$ for the Laplacian and Gaussian kernels. First, as either of them approaches zero, the optimal solution is simply the best sample cost.\footnote{This behavior for $\lambda$ is trivial and pointed out already in~\cite{rudi_finding_2024}. For $\sigma$, we have $\bm{K}\to\bm{I}$ as $\sigma \to 0 $. The feature vectors $\bm{\Phi}_i$ thus become the elementary vectors, and the optimal solution of~\eqref{eq:max-ksos} becomes $\bm{B}^\star=\mathrm{Diag}\left(f_1-c^\star, \ldots, f_N-c^\star\right)$.}
When $\sigma$ becomes too large, the kernel matrix approaches a constant (zero) matrix, meaning its effective dimension collapses, in which case the problem will become infeasible. Choosing $\lambda$ too large will not lead to infeasibility, but will also lead to high bias. These asymptotic properties can be verified in Fig.~\ref{fig:calibration}.

\noindent
\textbf{Other parameters.}
The initial center~$\bm{z}_0$ and radius~$r_0$ define the initial search space; we set them according to each problem's rough scale. The shrink factor $\gamma$ can be lower when the optimizer's quality is higher. 

In practice, we maximize~$N$ based on the computational budget, choose a nominal~$\sigma$ accordingly, and then tune both~$\sigma$ and~$\lambda$ using a calibration dataset.
We show an example for this calibration choosing the kernel type, with corresponding $\sigma$, and $\lambda$ in Sec.\ref{sec:implementation}. The optimal shrink factor $\gamma$ and number of restarts $w$ can be determined equivalently.

\subsection{Initializing local solvers}
While \kernelSOS{} can discover solutions across the entire search space, it can be less efficient than local solvers that exploit problem structure or derivative information.
Those methods, however, are prone to getting stuck in poor local minima.

A natural strategy is therefore to combine the two: to run \kernelSOS{} for a limited number of restarts to obtain a candidate near the global optimum, and then switch to a local solver to refine the solution.
This hybrid approach is particularly advantageous for problems that are too complex or high-dimensional for \kernelSOS{} to solve accurately in a reasonable time, but where the performance of local solvers is strongly dependent on the initial guess.
\section{Experiments}
\label{sec:problem}

\subsection{Example problems}\label{sec:problem-ro}
We assess the practical performance of \kernelSOS{} on the following two representative example problems.

\noindent
\textbf{Range-only localization}.
First, we consider the classic state-estimation task of \ac{RO} localization, which arises in robotics and positioning applications. It can be formulated as
\begin{equation}\label{eq:ro}
    \min_{\bm{x}} \sum_{i=1}^m \frac{1}{\sigma_i^2}{\left( d_i - \| \bm{x} - \bm{a}_i \|_2\right)}^2 =: f_{\text{RO-non-sq}}(\bm{x}), 
\end{equation}
where $\bm{x}\in\mathbb{R}^k$ is an agent's position in $k$ dimensions, $d_i$ are distance measurements to known anchors $\bm{a}_i\in\mathbb{R}^k$, $i \in [m]$, 
and $\sigma_i^2$ are chosen weights. The solution corresponds to the maximum-likelihood estimator under the assumption that the distance measurements are Gaussian distributed with mean $\|\bm{x}-\bm{a}_i\|_2$ and variance $\sigma_i^2$. To make $f_{\text{RO-non-sq}}$ polynomial for the standard \ac{SOS} method, two known remedies exist.

First, we can square the measurements $d_i$, in which case the cost function becomes a quartic polynomial, which we call $f_{\text{RO-sq}}$.
In this case, global solutions can be identified~\cite{beck_exact_2008,olsson_optimal_2016}. However, as we will show, this may lead to skewed results as it does not correspond to the assumed noise model.

Another approach introduces new unit-vector variables $\bm{n}_i$ and reformulates~\eqref{eq:ro} as follows~\cite{halsted_riemannian_2022}:
\begin{equation}\label{eq:ro-non-sq}
    \min_{\bm{x}, {\bm{n}_i\in \text{S}^k}} \sum_{i=1}^m \frac{1}{\sigma_i^2}\norm{\bm{n}_i d_i-(\bm{x}-\bm{a}_i)}_2^2,
\end{equation}
where $\text{S}^k$ denotes the unit sphere. One can show that the optimal $\hat{\bm{n}}_i$ are always aligned with $\bm{x} - \bm{a}_i$; the two formulations are thus equivalent. We will show in Section~\ref{sec:results-ro} that using this heuristic outperforms squaring the measurements in terms of accuracy, under the Gaussian noise model on distances. However, we need to introduce $mk$ new variables, and $m$ non-convex equality constraints. Thus, this approach scales poorly with problem dimension.

\noindent
\textbf{Trajectory optimization}.
Secondly, we consider the following instance of \ac{TO}:
\begin{equation}
    \label{eq:traj-opt}
    \begin{aligned}
        \min_{\bm{u}_{1:T}} \, & \| \bm{x}_{T+1}(\bm{u}_{1:T}) \|^2 + \rho \sum_{t=1}^T \| \bm{u}_t \|^2 =: f_{\text{TO}}\left(\bm{u}_{1:T}\,|\,\bm{x}_{\text{start}}\right) \\
        \text{s.t. } &\bm{x}_{t+1} = g(\bm{x}_t, \bm{u}_t),\, \bm{x}_1=\bm{x}_{\text{start}},
    \end{aligned}
\end{equation}
where, without loss of generality, the desired goal state is set to the origin, and $\bm{u}_{1:T}:=\{\bm{u}_1, \ldots, \bm{u}_T\}$ represents the control input sequence over the time window $T$. The initial state is $\bm{x}_{\text{start}}$ and the final state $\bm{x}_{T+1}$ is obtained through roll-outs of a given dynamics function $g$, placing us in a single-shooting setting. In this paper, rollouts are performed in a simulator, but they could be equivalently carried out on a real-world system. Note that we use a quadratic cost, which is a common choice in the literature, but this is not a requirement for \kernelSOS{}.
Such problems are typically much harder to reformulate as polynomial models, and motivate the use of methods that do not require closed-form models, such as \kernelSOS{}.

\subsection{Implementation details}
\label{sec:implementation}

\noindent\textbf{Custom interior-point solver.}
The main computational bottleneck is the solution of~\eqref{eq:max-ksos}. We implement the custom damped Newton algorithm detailed in~\cite{rudi_finding_2024}, which exploits the low-rank structure of the constraints,\footnote{Each constraint in~\eqref{eq:max-ksos} is of the form $b_i=\text{Tr}(\bm{B}\bm{A}_i)$ with $\bm{A}_i$ rank one.} and only involves iterations on the $N$ dual variables~$\alpha_i$. We set its accuracy to $\epsilon=10^{-3}$ , and the maximum number of steps to $100$.\footnote{Our implementation of \kernelSOS{}, along with the custom Newton solver, is available at \url{https://github.com/Simple-Robotics/ksos-tools}.}

\noindent\textbf{Choice of hyperparameters.}
We discuss in Section~\ref{sec:hyper} how to choose the kernel, number of samples per restart $N$ and restarts $w$, depending on the function to optimize. 
\begin{figure}[t]
    \includegraphics[width=\linewidth]{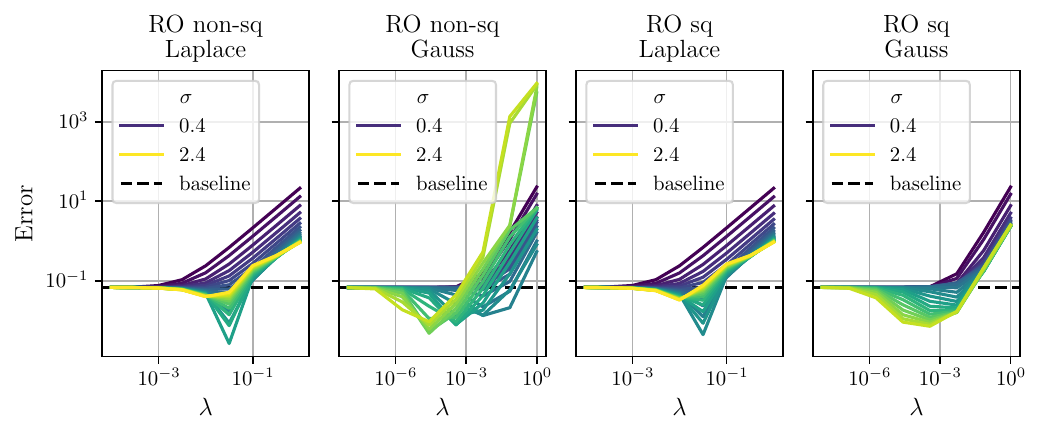}
    \caption{\textbf{Calibration results for range-only localization.} The localization error on a calibration dataset for different choices of kernel (Laplacian or Gaussian), kernel scale $\sigma$, and regularization parameter $\lambda$ is plotted. The baseline corresponds to picking the lowest-cost sample. While both kernel choices lead to similar best-case errors, the performance of the Gaussian kernel is less sensitive to the correct choice of $\lambda$.}
    \label{fig:calibration}
\end{figure}

\noindent\textbf{Number of samples.}
For the \ac{RO} problem, unless stated otherwise, we set the number of samples to $N=36$, and the number of restarts to $w=2$, at which point additional restarts showed diminishing returns. In calibration experiments, shown in Fig.~\ref{fig:calibration}, we choose the Gaussian kernel, which shows a more robust performance than the Laplacian, and set $\sigma=1.4$ and $\lambda=10^{-3}$ for the squared cost function, and $\sigma=1.0$ for the non-squared cost function.

For \ac{TO}, since the function to approximate is not as smooth, we choose the Laplace kernel. For experiments using \kernelSOS{} only, we combine results obtained for values of~$N$ ranging from $2$ to $500$, and $w$ ranging from $1$ to $10$; those results are aggregated and presented in terms of the total number of dynamics function calls $N\cdot w\cdot T$. When using \kernelSOS{} as an initialization scheme, we fix $N=100$ and the number of restarts to $w=2$.

\subsection{Range-only localization}\label{sec:results-ro}

We generate \ac{RO} simulations as follows. Over 20 random seeds, six known anchors and the target are sampled uniformly at random and rescaled such that their bounding box is $2\times 2$ units. Gaussian noise of varying scale is added to the ground-truth distances. We compare the following techniques: \emph{SampleSOS} solves the problem using feasible samples, as suggested by~\cite{lofberg_coefficients_2004}, for the unconstrained case with squared measurements, and by~\cite{cifuentes_sampling_2017} for the constrained case~\eqref{eq:ro-non-sq}.\footnote{Note that although sample-based, these algorithms assume a polynomial parametric form of known degree. In the unconstrained case, these algorithms are equivalent to \kernelSOS{} with a polynomial kernel.} 
By \textit{EquationSOS}, we denote the approach outlined, for example, in~\cite{beck_exact_2008,dumbgen_safe_2023}. 
We also compare to a naive baseline, which samples the search space on a uniform grid, and chooses the \textit{best sample} as the solution, and a \textit{local} solver initialized at 20 random points within the bounding box. 
Note that \kernelSOS{} can sample from the original formulation~\eqref{eq:ro}, which is considerably simpler than the reformulation~\eqref{eq:ro-non-sq}. 

\begin{figure}
    \centering
    \includegraphics[width=\linewidth,trim={0 0 1cm 0},clip]{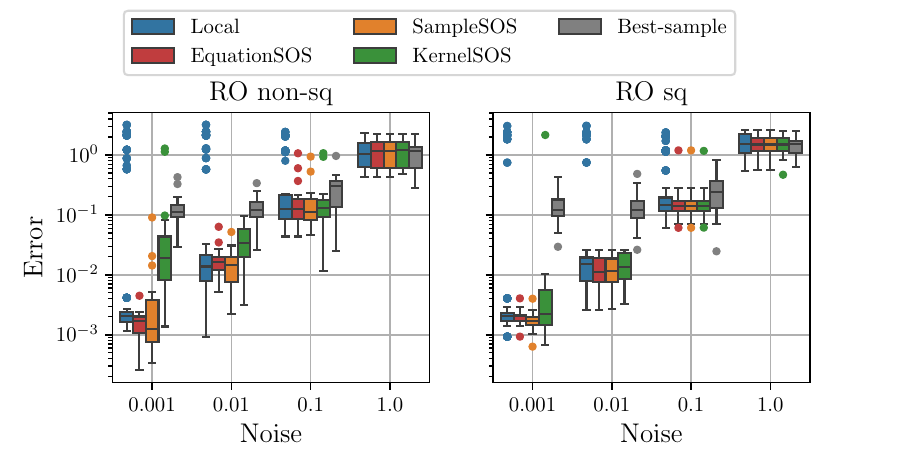}
    \includegraphics[width=\linewidth,trim={0 0 1cm 2.2cm},clip]{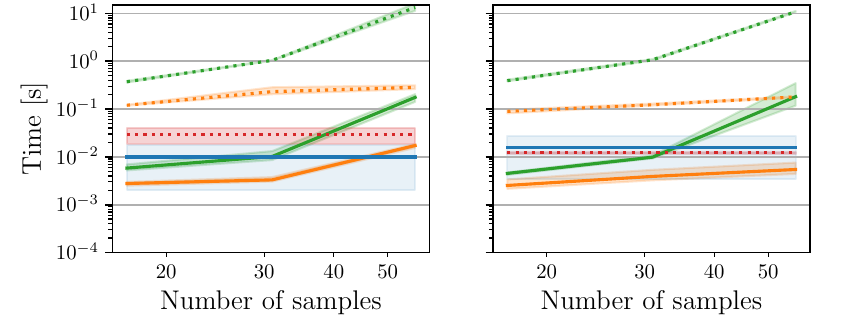}
    \caption{\textbf{Noise and time study for range-only localization.} Top: Distance to ground truth as a function of the noise level in range-only localization, using the non-squared (left) and squared (right) formulation. 
    \highlight{The \textit{local} solver often converges to local minima, while the (parametric) global \textit{EquationSOS} and \emph{SampleSOS} solvers perform consistently across noise levels and formulations.
    Notably, the (non-parametric) \kernelSOS{} method achieves similar performance for a wide range of noise levels, and significantly outperforms the best-sample baseline.
    Bottom: Run-time comparison. Dotted lines use the MOSEK solver, solid lines use custom solvers, in particular the open-source Newton solver.}}
    \label{fig:ro-noise-study}
\end{figure}

\begin{figure*}
    \vspace{5pt}
    \centering
    \begin{tabular}{c|c}
    RO non-sq & RO sq \\

\includegraphics[width=.45\linewidth,trim={1.9cm 0.3cm 1.9cm 1.1cm}, clip]{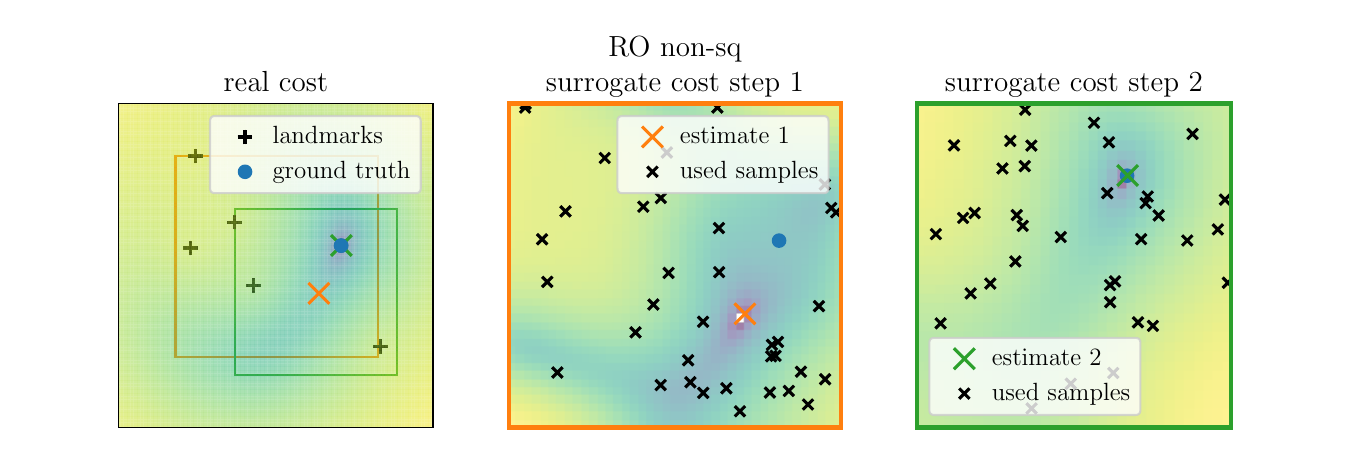} & \includegraphics[width=.45\linewidth,trim={1.9cm 0.3cm 1.9cm 1.1cm}, clip]{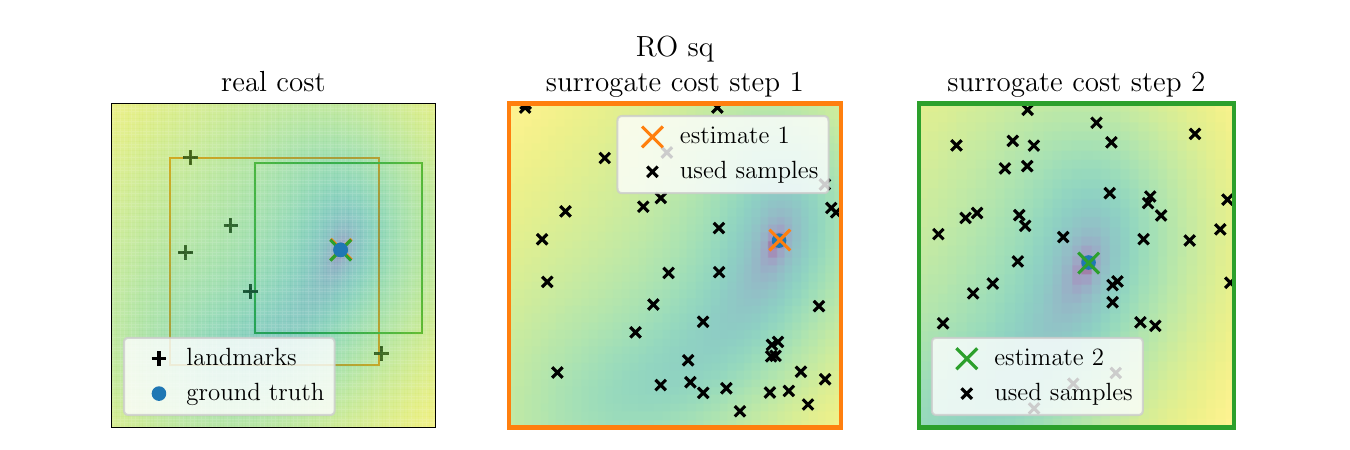}
    \end{tabular}
    \caption{\textbf{Illustration of the surrogate cost and the benefit of restarts.} The cost of the range-only problem is computed either using non-squared distances (\textbf{left half}) or squared distances (\textbf{right half}). In each block, we have on the left: real cost; center: surrogate function found by the first \kernelSOS{} step; right: surrogate function found by the first restart (second step). 
    The sampling regions of the first and second steps are highlighted with orange and green squares. 
    Black crosses represent the known landmarks, black x-marks represent the used samples. Orange and green x-marks denote the minimizers found by \kernelSOS{} initially and after restarting, respectively.}
    \label{fig:ro-nonsq-domain}
    \vspace{-5pt}
\end{figure*}

The results are provided in Fig.~\ref{fig:ro-noise-study}. We note that the local solver, while having the lowest mean error, exhibits many high-error outliers corresponding to poor local minima. This motivates the use of global solvers, as relying on such bad solutions, even if relatively rare, may lead to unsafe behavior. Second, we can see that the non-squared formulation results in significantly better accuracy. Considering the globally optimal solutions of the local solver for each individual setup, the error using the squared cost function is 40\% (average) and 17\% (median) higher relative to the non-squared cost function, averaging across all noise levels. 

The globally optimal solvers, \textit{EquationSOS} and \textit{SampleSOS}, achieve on average similar accuracy (since they solve equivalent problem formulations), but they require specific knowledge of the problem. In contrast, \kernelSOS{} achieves competitive performance, in particular at high noise levels, without access to a parametric cost. It is also notable that \kernelSOS{} clearly outperforms the best-sample baseline, showing that the interpolation based on the chosen kernel helps retrieve a better solution. 

In terms of runtime, shown in the bottom of Fig.~\ref{fig:ro-noise-study},~\kernelSOS{} has similar empirical complexity as \textit{SampleSOS}, since both methods solve an \ac{SDP} of the same structure. For \textit{SampleSOS}, the kernel is always singular, and we only keep the non-zero basis functions, which reduces the problem size and leads to faster solve times. This is, however, only possible because of the polynomial reformulations. For both cases, we see a large improvement when using the custom low-rank exploiting Newton solver compared with MOSEK, which is notable since the custom solver uses non-optimized \textit{Python} code. 
Using the custom solver, \kernelSOS{} is even competitive with \textit{EquationSOS}; in particular, on the non-squared formulation. 
The sample-based formulations are expected to become even more competitive when \textit{EquationSOS} needs a high degree in the moment hierarchy for tightness, which is not the case for this particular example problem. 

In Fig.~\ref{fig:ro-nonsq-domain}, we visualize the surrogate functions for \ac{RO} localization, across restarts. We can observe qualitatively that \kernelSOS{} implicitly fits a relatively accurate surrogate function. 
For the non-squared setting, the first step of \kernelSOS{} identifies the correct region to zoom in on, while the second restart is necessary to achieve higher accuracy. For the squared setting, the cost is less localized, and a good solution is already found with the first iteration. 

\subsection{Zeroth-order trajectory optimization}
While \textit{SampleSOS} proves to be less computationally expensive in settings where a polynomial model of the problem is available, we now show that \kernelSOS{} can be used in settings without exploiting polynomial, let alone parametric, models. 
We evaluate the performance of \kernelSOS{} for \ac{TO} on the swing-up task for single and double pendulum systems, simulated using the Pinocchio~\cite{carpentier2019pinocchio,pinocchioweb} library. 
We compare \kernelSOS~to \ac{CMAES}, as a fair comparison with \ac{BO} would require integrating an active sampling strategy into \kernelSOS{}, which we leave for future work.
\begin{figure}[tb]
    \centering
    \includegraphics[width=\linewidth]{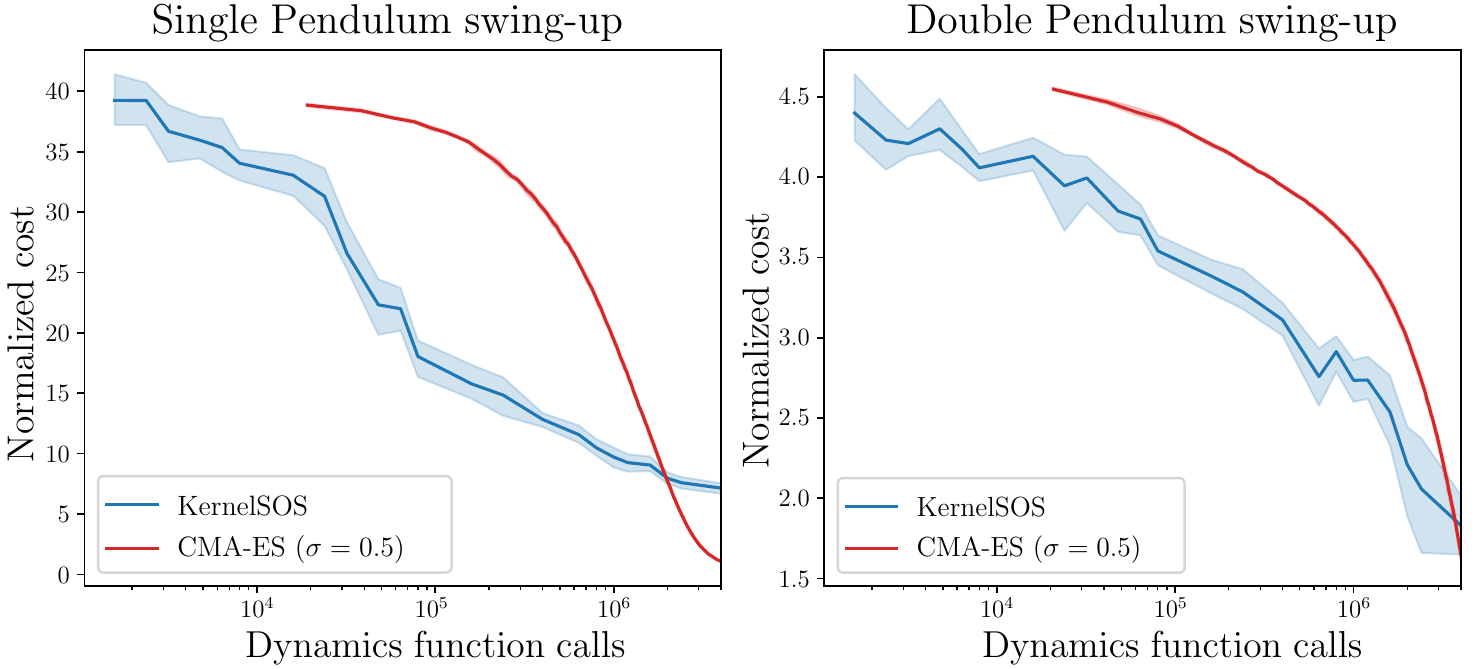}
    \caption{\textbf{\kernelSOS{} \emph{vs.}~CMA-ES in trajectory optimization.} In both cases, the horizon $T$ is set to $4/0.005$. \kernelSOS{} results are averaged over 10 runs of the algorithm to evaluate the influence of the random samples. The shaded region represents the standard deviation for the randomness of the method.}
    \label{fig:TO-plots}
    \vspace{5pt}
\end{figure}
In Fig.~\ref{fig:TO-plots}, we compare the normalized cost of trajectories found by \kernelSOS{} and \ac{CMAES}, as a function of the number of calls to the simulator. We can see that \kernelSOS{} outperforms \ac{CMAES} for a low number of samples, while they approximately match for a higher number of samples. This shows that the global nature of \kernelSOS{} allows it to find good trajectories from the entire space from the start, while \ac{CMAES} tends to explore the search space more incrementally. 

Both methods find trajectories that swing up the pendulum, but the accuracy of the solution may be insufficient. 
The ability to find relatively good solutions with a small number of samples motivates the use of \kernelSOS{} as an initialization method for local optimization methods, as discussed next.

\subsection{Trajectory optimization including local solvers}

\begin{figure}[htb]
    \centering
        \centering
        \includegraphics[width=\linewidth]
        {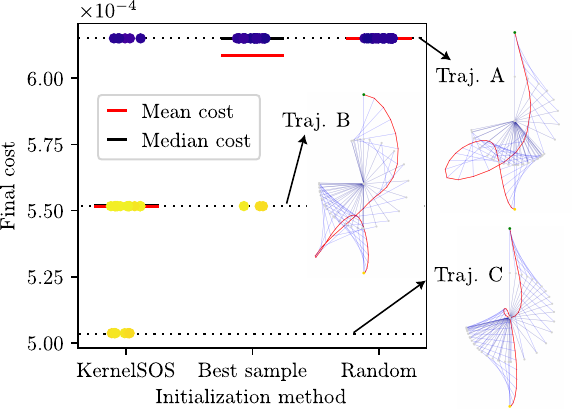}
    \caption{\textbf{Solution quality of local solver.} 
    We compare the performance of FDDP for the double-pendulum swing-up task, depending on the initialization. Dots are colored from yellow to blue depending on the number of iterations needed for FDDP to converge after initialization, with yellow indicating fewer iterations. 
    Using \kernelSOS{} as an initializer reduces the cost by 12\% on average compared to a random initialization. 
    In particular, we observe the emergence of three discrete modes at three distinct cost levels, corresponding to trajectories A, B, and C. Notably, the best trajectory (Trajectory C) is only found thanks to the \kernelSOS{} initialization.}
    \label{fig:warmstarting-combined}
\end{figure}

We compare the performance of \ac{FDDP}~\cite{fddp}, implemented in the Aligator library~\cite{jallet2025proxddp,aligatorweb}, for different initialization mechanisms. The algorithm stops once the primal feasibility falls below $10^{-8}$. We use the double pendulum swing-up task as a benchmark, and compare the following initialization methods: \textit{random}: a single random input sequence is sampled from the search space; \textit{best-sample}: 100 random input sequences are sampled from the search space, and the one with the lowest cost is used as an initialization.

Cost-performance results for 30 trajectories per initialization method are shown in Fig.~\ref{fig:warmstarting-combined}. We observe that using \kernelSOS{} as an initialization reduces the cost by 12\% on average compared to a random initialization, and by 10\% compared to the one using the best sample. Due to the problem's high dimensionality, the best-sample initialization yields only a small improvement over random initialization, whereas \kernelSOS{} significantly improves the initialization. Indeed, although the problem has only two degrees of freedom, planning a trajectory requires optimizing the control inputs at each time step, resulting in a high-dimensional search space (here, of size 400). In such a setup, the ability of \kernelSOS{} to interpolate the cost function based on a limited number of samples is crucial for finding a good initialization.

It is interesting to note that the trajectories found after running \ac{FDDP} always converge to one of three distinct costs, which correspond to three local minima of the problem. These three modes are visualized in Fig.~\ref{fig:warmstarting-combined} on the right. Notably, \kernelSOS{} is the only initialization method after which the lowest-cost solution can be found.

\begin{figure}
    \vspace{5pt}
    \centering
    \includegraphics[width=\linewidth,trim={0 0.1em 0 0}, clip]{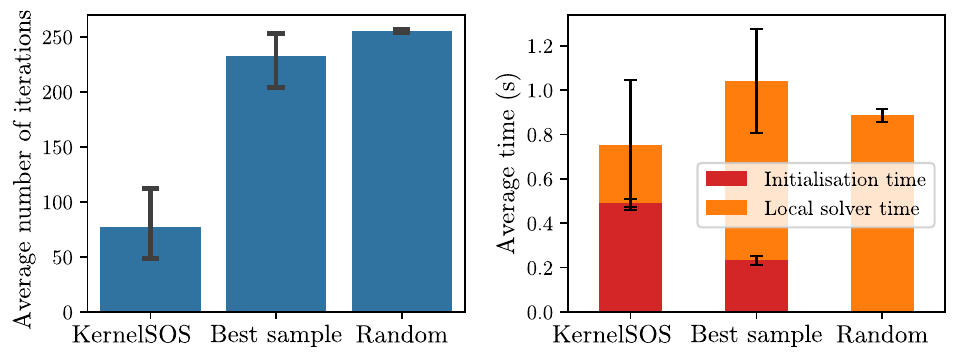}
    \caption{\textbf{Overall convergence and runtime comparison.} \kernelSOS{} initialization reduces the number of iterations by 77\% on average, leading to a similar total time as random and best-sample initialization despite the time required for \kernelSOS{} to solve.}
    \label{fig:warmstarting-time}
    \vspace{-5pt}
\end{figure}

\begin{figure}
    \vspace{5pt}
    \centering
    \includegraphics[width=\linewidth, trim={0 0.0em 0 0.3em}, clip]{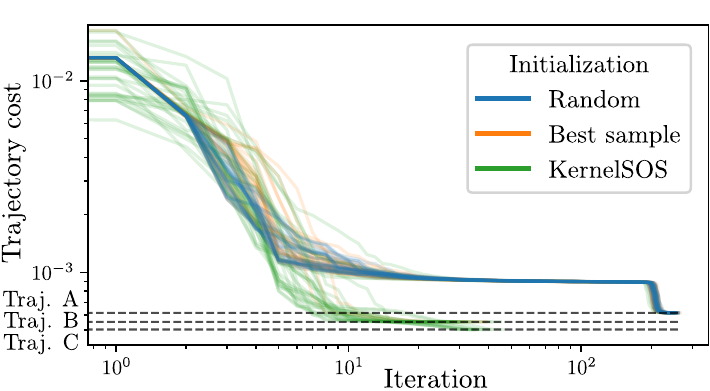}
    \caption{\textbf{Cost evolution during the FDDP optimization phase.} Dashed lines represent the costs of the three discrete modes corresponding to the trajectories graphed in Fig.~\ref{fig:warmstarting-combined}. On average, we observe improved convergence to lower costs in the case of \kernelSOS{}, even though the initial costs are not significantly lower.}
    \label{fig:trajectories-convergence}
    \vspace{-5pt}
\end{figure}

Finally, we investigate whether the additional time taken by the initialization step is worth the improvement in performance. Counterintuitively, we observe that the total time taken when initializing with \kernelSOS{} is slightly smaller than the time taken when randomly initializing, despite the initialization part being considerably more costly for \kernelSOS{}. This is because \kernelSOS{} initialization reduces the number of iterations needed by \ac{FDDP} to converge by 77\% on average, as shown in Fig.~\ref{fig:warmstarting-time}. As the initial guess provided by \kernelSOS{} is closer to a local minimum or in a region with better conditioning, \ac{FDDP} requires fewer iterations.  This can be seen in Fig.~\ref{fig:trajectories-convergence}, where \kernelSOS{} makes FDDP converge faster and to lower values than other initialization methods. On the other hand, the best-sample initialization reduces the number of iterations needed to converge only slightly, while requiring a significant time to evaluate the cost of 100 trajectories; as a result, the best-sample method is the slowest of the three. In summary, \kernelSOS{} discovers higher-quality solutions slightly faster than random sampling methods.
\section{Conclusion and Future Work}\label{sec:conlusion}

We enabled the successful application of \kernelSOS{}, a strong new contender in the global optimization field, for optimal control and estimation problems. By relying only on samples, it generalizes the \ac{SOS} framework to non-polynomial, non-parametric problems and allows it to include black-box elements such as high-fidelity simulators. Our contributions lie in bridging theory and practice: we identified and addressed the key ingredients that make \kernelSOS{} usable in robotics and control, including strategies for restarts, hyperparameter calibration, and recovering minimizers. We demonstrated its use as a standalone solver and as an initialization method, thereby combining the global search capabilities of \kernelSOS{} with the efficiency and accuracy of local optimization methods.

In future work, the method could be extended with improved sampling strategies, such as active sampling as in Bayesian optimization, to make exploration more efficient. 
Adding the ability to handle constraints would make the method applicable to an even broader class of control problems. On the application side, the role of \kernelSOS{} as an initialization method merits further investigation, particularly for policy learning, where it could improve search directions, accelerate convergence, and enhance sample efficiency.


\section*{Acknowledgments}
We warmly thank Alessandro Rudi, Francis Bach, Corbinian Schlosser, and Lorenzo Shaikewitz for many insightful discussions and comments on the manuscript.

\printbibliography

\appendices
\section{Finding the minimizer via the dual problem}\label{sec:appendix-dual}

In this section, we provide intuition for the solution candidate given by~\eqref{eq:sol}. The dual problem of~\eqref{eq:max-ksos} is~\cite{boyd_convex_2004}:
\begin{equation}\begin{aligned}
\min_{\bm{\alpha}, \bm{M}} \,\, & \sum_i f(\bm{x}_i) \alpha_i \\
\text{s.t. } & \sum_i \alpha_i = 1, \,\,
 \bm{M}:=\sum_i \alpha_i \bm{\Phi}_i\bm{\Phi}_i^\top+\lambda \bm{I}\succeq 0,
\end{aligned}\label{eq:dual}
\end{equation}
where $\bm{M}$ is the dual variable associated with $\bm{B}\succeq 0$, 
and $\bm{\alpha}\in\mathbb{R}^N$ contains the dual variables associated with the equality constraints. In this context, we understand the $\bm{\Phi}_i$ vectors as $\bm{\Phi}_i=\bm{\phi}(\bm{x}_i)$ for a feature function $\bm{\phi}$, where $\bm{x}_i$ is the $i$-th sampling point.

When we have strong duality and a unique minimizer, the dual problem, as $N\to\infty$, has a rank-one optimal solution of the form $\bm{M}^\star=\int \bm{\phi}(\bm{x})\bm{\phi}(\bm{x})^\top \mathrm d\mu^\star(\bm{x})=\bm{\Phi}^\star{\bm{\Phi}^\star}^\top$, where $\mu^\star$ is the optimal measure --- a Dirac at the optimal solution $\bm{x}^\star$~\cite{henrion_moment-sos_2020}. 
The identity~\eqref{eq:sol} emerges from $\bm{M}^\star=\sum_i \alpha_i^* \bm{\Phi}_i\bm{\Phi}_i^\top + \lambda\bm{I}$ by~\eqref{eq:dual}. To obtain a candidate minimizer, we let $\lambda$ tend to zero. Then, we assume that there exists a matrix $\bm{A}$ such that $\trace{\bm{A},\bm{\phi}(\bm{x})\bm{\phi}(\bm{x})^\top}=\bm{x}$ on $\Omega$.\footnote{Note that the identity function $\bm{x}$ may in general not be in the \ac{RKHS} defined by the chosen kernel. However, for a bounded set $\Omega$, the function $\bm{x}$ can be arbitrarily well approximated in the \ac{RKHS}, provided it is dense in $L^2(\Omega)$, as is the case for, e.g., the Gaussian or Laplace kernels.} Using this matrix, we obtain:
\begin{equation}
\begin{aligned}
    \bm{x}^\star = \trace{\bm{A}, \bm{\Phi}^\star{\bm{\Phi}^\star}^\top} &= \trace{\bm{A}, \sum_i \alpha_i^\star \bm{\Phi}_i\bm{\Phi}_i^\top} \\ 
    &= \sum_i \alpha_i^\star \trace{\bm{A}, \bm{\Phi}_i\bm{\Phi}_i^\top} = \sum_i \alpha_i^\star \bm{x}_i.
    \end{aligned}
\end{equation}
Note that when there are multiple global minima, the estimate may, in theory, be poor. We did not find this to be an issue for the considered example problems.

\end{document}